\documentclass[conference]{IEEEtran} 
\IEEEoverridecommandlockouts                
\usepackage{graphicx}
\usepackage{amsmath}
\usepackage{amssymb}
\usepackage{booktabs}
\usepackage{algpseudocode}
\usepackage{algorithm}
\usepackage{subcaption}

\algblockdefx[Name]{Repeat}{Until}{\textbf{repeat}}{\textbf{until}~}

\usepackage[pagebackref,breaklinks,colorlinks]{hyperref}
\providecommand{\keywords}[1]{\textbf{\textit{Index terms---}} #1}
\graphicspath{{figures/}}

\begin{document}

\title{Self-Adaptive Robust Motion Planning for High DoF Robot Manipulator using Deep MPC}

\author{Ye Zhang$^{1,*}$, Kangtong Mo$^1$, Fangzhou Shen$^1$, Xuanzhen Xu$^1$, Xingyu Zhang$^2$, Jiayue Yu$^2$ and Chang Yu$^2$\\
$^{1,*}$University of Pittsburgh, PA 15213, USA\\
$^1$University of Illinois Urbana-Champaign, IL 61820, USA\\
$^1$San Jose State University at San Jose, CA 95192, USA\\
$^1$Snap Inc. at Seattle, WA 98121, USA\\
$^2$George Washington University, DC 20052, USA\\
$^2$Warner Bro. Discovery at Culver City, CA 90232, USA\\
$^2$Northeastern University at Boston, MA 02115, USA\\
\centerline{$^{1,*}$yez12@pitt.edu, $^1$mokangtong@gmail.com, $^1$fangzhou.shen0922@gmail.com}\\ \centerline{$^1$xuanzhenxu@gmail.com, $^2$xingyu\_zhang@gwmail.gwu.edu, $^2$jiy048@gmail.com, $^2$chang.yu@northeastern.edu}
}
\maketitle

\begin{abstract}
In contemporary control theory, self-adaptive methodologies are highly esteemed for their inherent flexibility and robustness in managing modeling uncertainties. Particularly, robust adaptive control stands out owing to its potent capability of leveraging robust optimization algorithms to approximate cost functions and relax the stringent constraints often associated with conventional self-adaptive control paradigms. Deep learning methods, characterized by their extensive layered architecture, offer significantly enhanced approximation prowess. Notwithstanding, the implementation of deep learning is replete with challenges, particularly the phenomena of vanishing and exploding gradients encountered during the training process. This paper introduces a self-adaptive control scheme integrating a deep MPC, governed by an innovative weight update law designed to mitigate the vanishing and exploding gradient predicament by employing the gradient sign exclusively. The proffered controller is a self-adaptive dynamic inversion mechanism, integrating an augmented state observer within an auxiliary estimation circuit to enhance the training phase. This approach enables the deep MPC to learn the entire plant model in real-time and the efficacy of the controller is demonstrated through simulations involving a high-DoF robot manipulator, wherein the controller adeptly learns the nonlinear plant dynamics expeditiously and exhibits commendable performance in the motion planning task.
\end{abstract} 


\section{Introduction}
\label{sec:intro}
In the past decade, adaptive control strategies have gained considerable prominence within control theory and robotics application~\cite{Gao2024DecentralizedAA,liu2024adaptive100,10309449,gao2023autonomous}. Their capacity to rectify modeling inaccuracies and mitigate unforeseen disturbances has rendered them highly sought after as more intelligent and resilient control methodologies are pursued. A particularly potent and widely embraced adaptive control variant involves deploying a deep learning method~\cite{wu2023creating,zou2023joint,zhibin2019labeled,shen2024localization,tong2024robust,wu2023enhanced}. With their remarkable ability to approximate nearly any function within an acceptable margin of error, these methods have been extensively applied beyond control theory, notably in data science, for managing large and intricate data sets. Within traditional control theory, deep learning primarily functions as controller estimation mechanisms, enabling real-time learning of uncertainties in plant dynamics or disturbances~\cite{jin2024learning}, thereby progressively enhancing controller performance. For instance, in the works of~\cite{nubert2020safe}, adaptive dynamic inversion controllers were formulated using a neural network state observer—referred to as a modified state observer—within an auxiliary estimation loop to accommodate modeling uncertainties or unknown disturbances in non-affine~\cite{jin2024apeer}, non-square nonlinear systems~\cite{padhi2007model} extended this control scheme into an event-triggered control framework.

\begin{figure}[!t]
\centering
\begin{center}
\includegraphics[width=0.97\columnwidth]{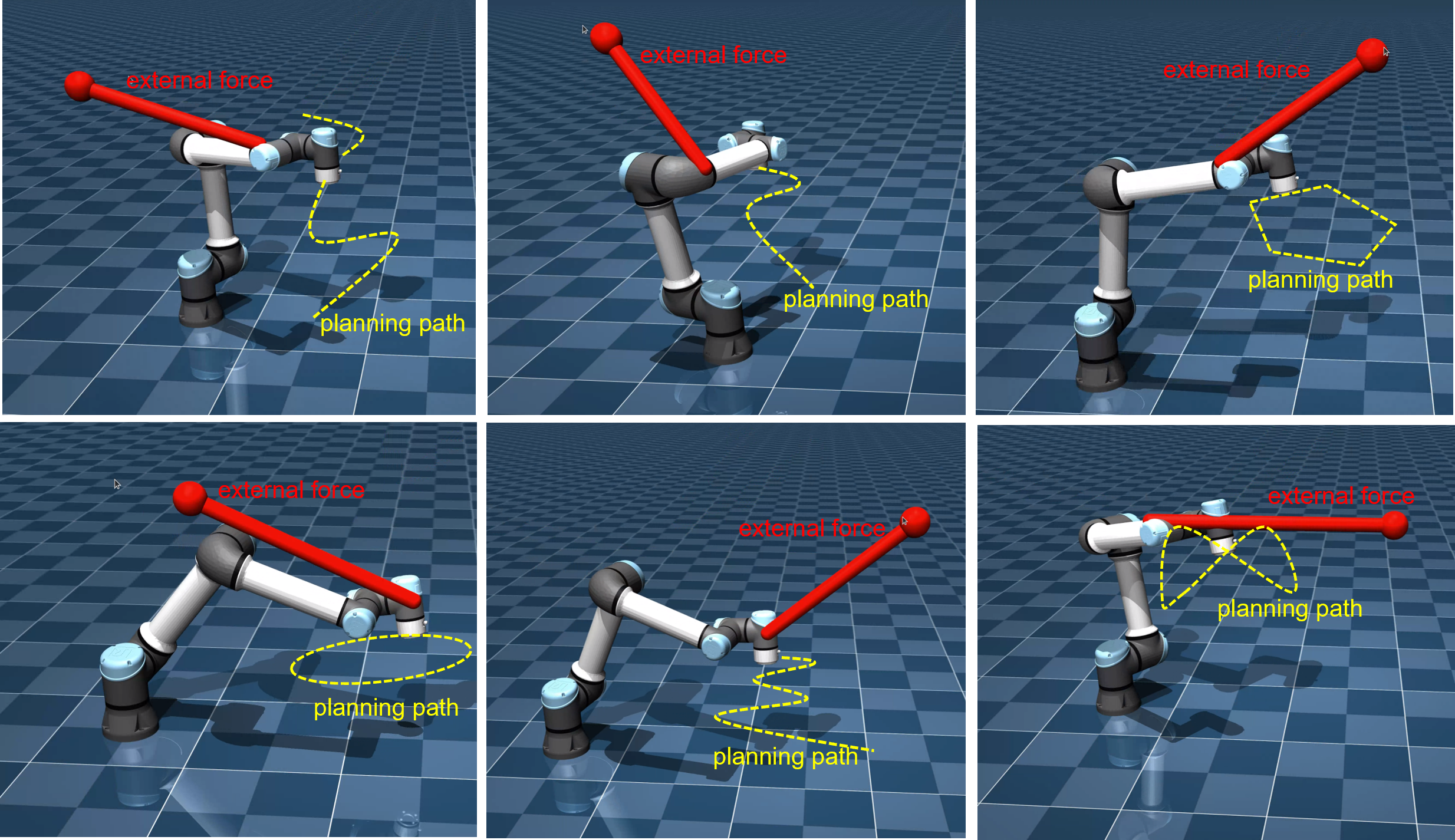}
\end{center}
\vspace{-2mm}
 \caption{Different motion planning tasks for a high-DoF robot arm(UR5) with deep MPC under different external forces applied on different parts to examine the self-adaptive performance.}
 \label{fig:overview}
 \vspace{-4mm}
\end{figure}

The amalgamation of MPC with neural networks leverages the predictive capabilities of MPC while enhancing its adaptive nature through neural network-based learning. This synergy allows for real-time adjustments to control strategies based on the continuous influx of new data~\cite{deng2024neslam}, thereby maintaining optimal performance despite uncertainties and external perturbations. The ability to anticipate future states and control inputs within MPC, combined with the adaptive learning prowess of neural networks, culminates in a sophisticated control methodology that is both robust and highly adaptable to dynamic environments.

Recent strides in deep learning have culminated in the advancement of MPC, which stands as more potent and intrinsically more intricate iterations of the classical control method, typically characterized by substantially larger architectures. A thorough analysis of MPC, encompassing their evolution, predominant models~\cite{zou2022unified,huang2024research}, applications, and prospective research avenues, is presented in~\cite{afram2014theory,liu2023financial}. Within the sphere of control theory, MPC are extensively harnessed for reinforcement learning applications~\cite{tong2022adaptive,deng2024incremental}. The deep reinforcement learning framework is comprehensively delineated in~\cite{abu2022deep}, accompanied by discussions on its practical applications and extant implementations.

In another investigation, combine MPC and reinforcement learning to regulate a quadcopter drone in ~\cite{salzmann2023real} show successful implementation and high performance. Furthermore,~\cite{saviolo2022physics} explored MPC with DRL for UAV trajectory tracking, employing a baseline controller that concurrently performs initial control while training the deep MPC controller~\cite{wang2024research}. Upon the MPC controller achieving sufficient policy generation\cite{wu2023cargo}, control is transitioned to the traditional controller~\cite{zhang2020manipulator}, with online training continuing for varied trajectories not included in the initial training set. However, few studies focus on self-adaptive control with deep learning methods on motion planning tasks for high-DoF robot manipulation with unknown environment disturbance.

This paper focuses on combining the reinforcement learning method with MPC for high-DOF robot manipulators in motion planning tasks with external disturbance on its different parts. The contributions of the paper are:
\begin{itemize}
    \item we develop a novel deep MPC algorithm concerning the unknown disturbance from external environment. The robustness of our method are examine through our simulation work.
    \item Our deep MPC method shows high performance and strong robustness not only for the high-DoF robot manipulator, but also in other robot platform. We examine our method on different types of robot in simsultion environment, which shows strong extensity.
\end{itemize}

\section{Methodology}
\label{sec:method}
Consider the dynamics of an uncertain, nonlinear system:
\begin{equation}
    \dot{x}_k = f(x_k, u_k) + Bu_k+\xi(u_k)
\end{equation}
where $f(x_k)$ is the nonlinear, known system dynamics, B is the control matrix, $u_k$ is the control signal, and $d(x_k)$ is the state-dependent uncertainty. The system can be simplified by distributing $B$:
\begin{equation}
    \dot{x}_k = f(x_k) + \check{f}(x_k) + Bu_k
\end{equation}
\begin{equation*}
    \check{f}(x_k) = B\xi(x_k)
\end{equation*}
The objective of performance is to maneuver the uncertain system towards a target system characterized by the specified dynamics is:
\begin{equation}
    \dot{x}^{d}_k = f^{*}(x^d_k,u_L)
\end{equation}
where $u_{L}$ is some nominal control signal. To achieve this, we utilize a dynamic inversion controller to substitute the original system's dynamics with the desired ones. Nevertheless, a notable drawback of dynamic inversion controllers is their susceptibility to modeling inaccuracies in the system dynamics; consequently, the inherent uncertainty of the plant presents a substantial challenge. To remedy this, past literature~\cite{8431513} have relied on learning the uncertainty using a modified state observer.

\begin{figure*}[t]
\centering
\begin{subfigure}{2.3in}
    \includegraphics[width=\textwidth]{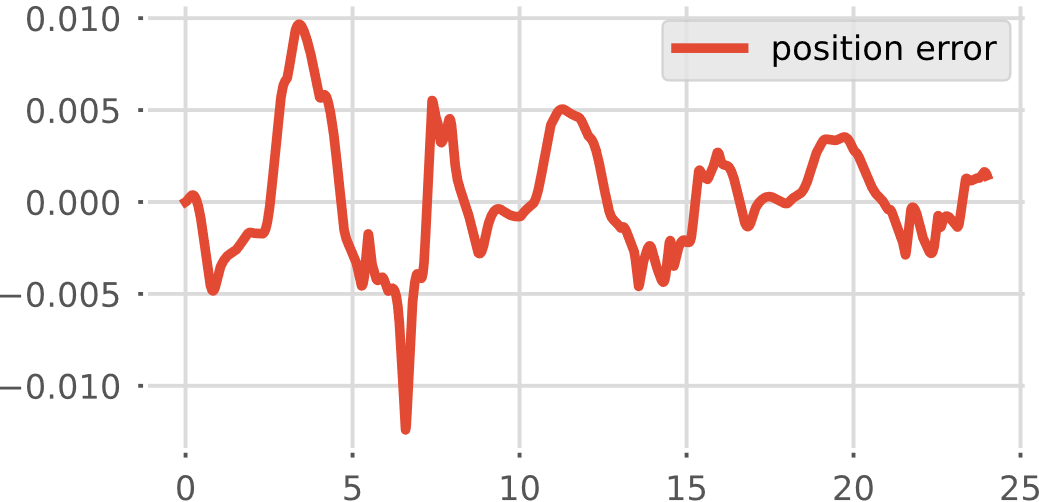}
    \caption{}
    \label{sub_a}
\end{subfigure}
\begin{subfigure}{2.3in}
    \includegraphics[width=\textwidth]{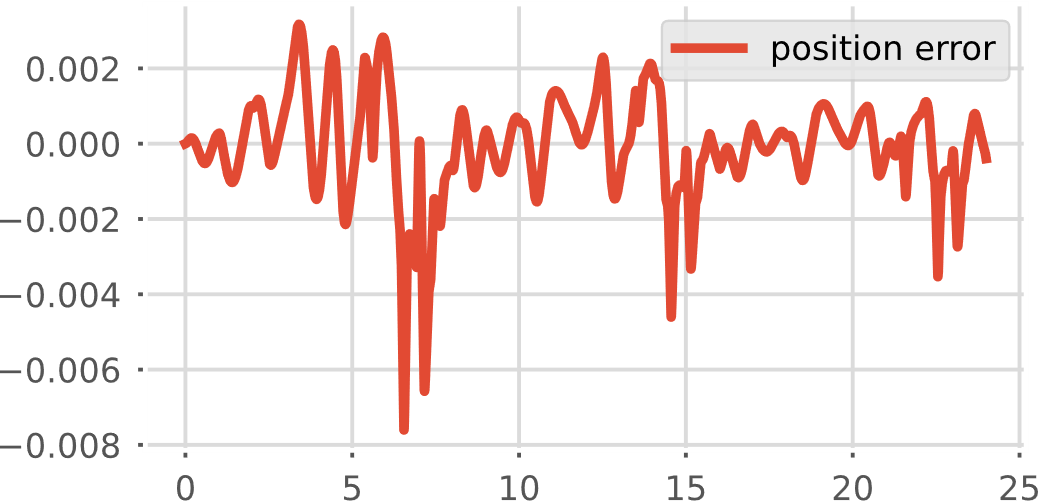}
    \caption{}
    \label{sub_b}
\end{subfigure}
\begin{subfigure}{2.3in}
    \includegraphics[width=\textwidth]{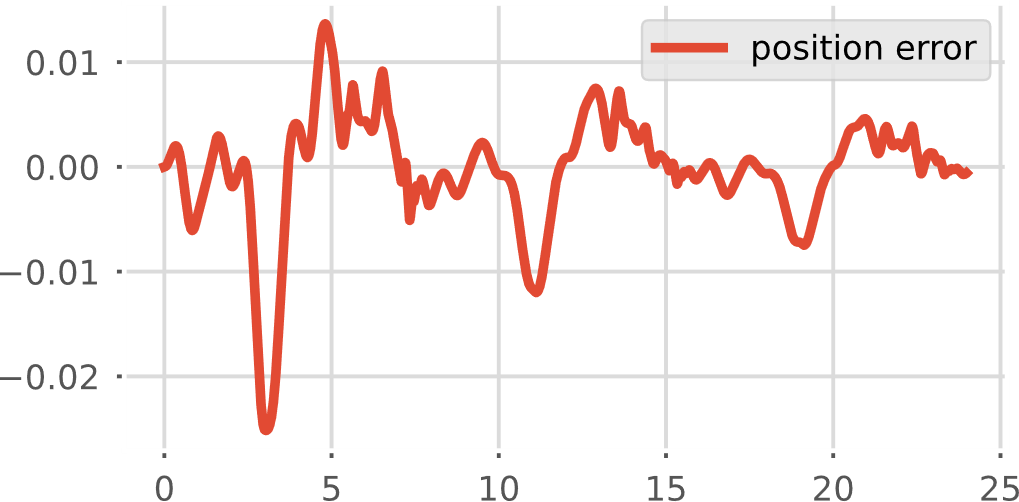}
    \caption{}
    \label{sub_c}
\end{subfigure}
\begin{subfigure}{2.3in}
    \includegraphics[width=\textwidth]{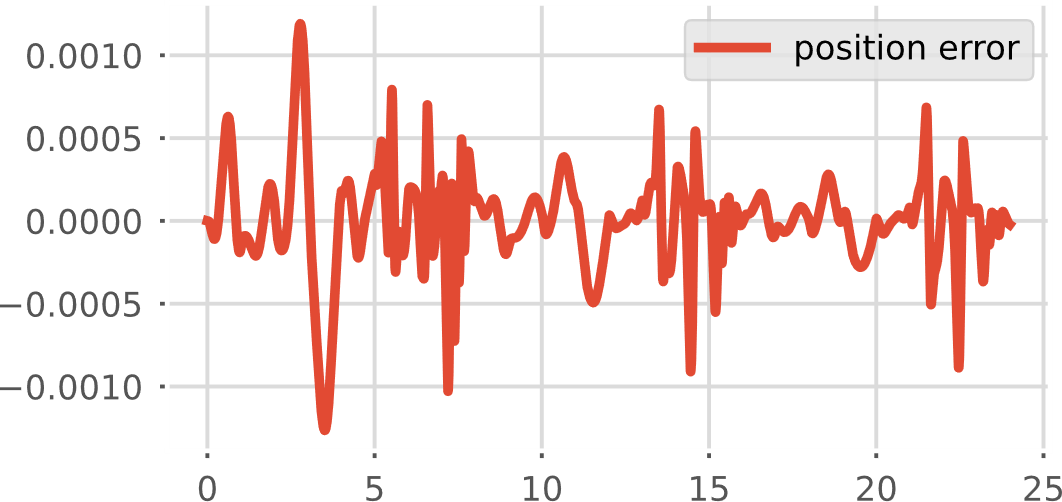}
    \caption{}
    \label{sub_d}
\end{subfigure}
\begin{subfigure}{2.3in}
    \includegraphics[width=\textwidth]{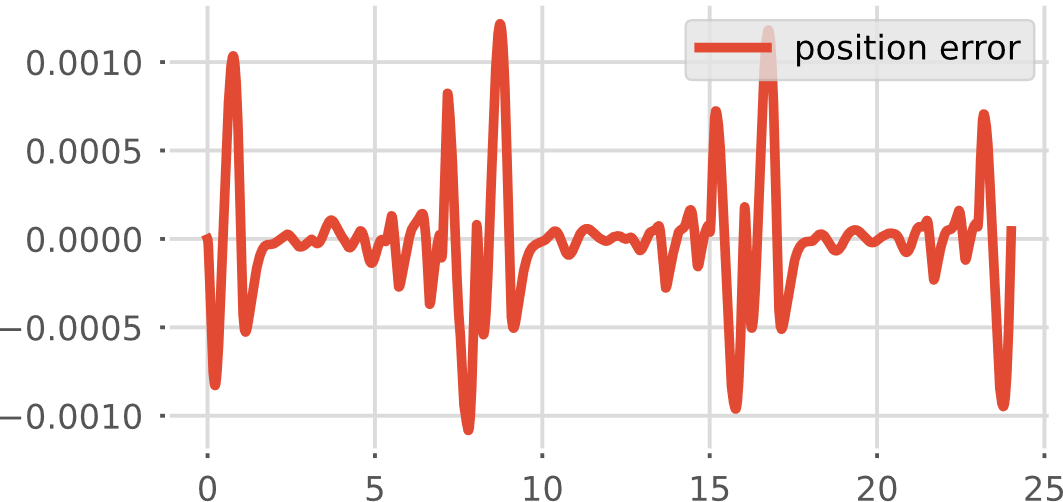}
    \caption{}
    \label{sub_e}
\end{subfigure}
\begin{subfigure}{2.3in}
    \includegraphics[width=\textwidth]{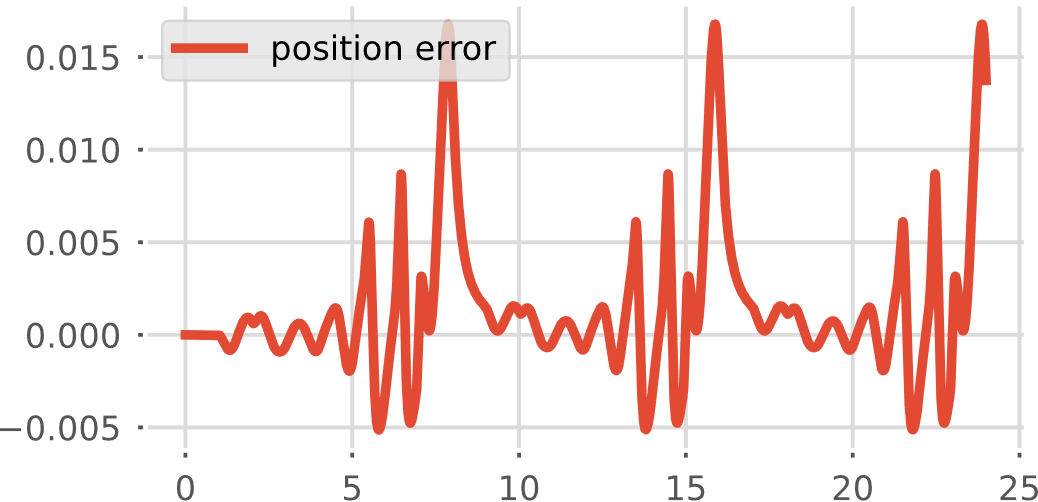}
    \caption{}
    \label{sub_f}
\end{subfigure}
\caption{Tracking errors for the end-effector of the UR5 robot arm in different path planning tasks. From (a) to (f) denotes the tracking error concerning the scenarios Fig.\ref{fig:overview}}
\label{results}
\end{figure*}

Given that the primary objective of the controller is to steer the actual system state $ x_k $ towards the desired system state $ x^d_k $, we examine the following equation:
\begin{equation}\label{contForm1}
    \dot{x}_k - \dot{x}^{d}_k + \Gamma(x_k-x^{d}_k) = 0
\end{equation}
which ensures first-order asymptotically stable error dynamics \cite{krener1994nonlinear}. It is important to note that $\Gamma$ is a user-defined, positive-definite gain matrix. By substituting the system dynamics into equation (\ref{contForm1}):
\begin{equation}
     \check{f}(x_k) + Bu_k - f^{*}(x^{d}_k,u_L) + \Gamma(x_k-x^{d}_k)=0
\end{equation}
Let's reorganize above and then we can get:
\begin{equation}
    Bu_k = f^{*}(x^{d}_k,u_L) + \check{f}(x_k))-\Gamma(x_k-x^{d}_k)
\end{equation}
Assuming that the uncertainty estimate, $\hat{f}(x_k)$, is available and $\hat{f}(x_k) \approx{f(x_k) + \check{f}(x_k)}$, this can be substituted for the system dynamics as:
\begin{equation}
    Bu_k = f^{*}(x^{d}_k,u_L)-\hat{f}(x_k)-\Gamma(x_k-x^{d}_k)
\end{equation}
Then, if $B$ is invertible, the dynamic inversion controller can be completed by multiplying by the inverse as:
\begin{equation}
    u_k =B^{-1}(f^{*}(x^{d}_k,u_L)-\hat{f}(x_k)-\Gamma(x_k-x^{d}_k))
\end{equation}
If matrix $ B $ is non-invertible, slack variables can be seamlessly integrated~\cite{zou2023capacity}. This involves augmenting the formulation by introducing and subtracting the slack matrix $ B_{s} $ and the slack control signal $ u_{s} $, thus transforming the control matrix into an invertible configuration~\cite{yu2024stochastic}.
First, the tracking error, estimation error, and estimate tracking error are defined as follows:
\begin{align}
    e_{r} &\triangleq{x_k - x^{d}_k}
\end{align}
The tracking error can now be expressed as a combination of the estimation error and the estimate tracking error as follows:
\begin{equation*}
    e_{r} = x_k - \hat{x}_k+\hat{X}_k -x^{d}_k
\end{equation*}
\begin{equation}\label{errDy1}
    e_{r}=e_{a}+\hat{e}_{r}
\end{equation}

Here let's expand the estimate tracking error derivative as:
\begin{equation*}
    \begin{array}{lcl}
    \dot{\hat{e}}_{r}= \hat{f}(x_k)+Bu_k -\Lambda(x_k-\hat{x}_k)
    \end{array}
\end{equation*}

Now using $Bu_k = f^{*}(x^{d}_k,u_L)-\hat{f}(x_k)-\Gamma(x_k-x^{d}_k)$:

\begin{equation}
    \dot{\hat{e}}_{r}= -(\Gamma(x_k-x^{d}_k)+\Lambda(x_k-\hat{X}_k))
\end{equation}

Finally, we can get:
\begin{equation}
    \dot{e}_{r}=\tilde{f}(x_k) - \Gamma(x_k-x^{d}_k)
\end{equation}

The proposed method integrates concepts from~\cite{krener1994nonlinear}, extending the framework to accommodate robust scenarios.
The following optimization problem characterizes the proposed Robust Model Predictive Control scheme for setpoint tracking~\cite{zou2023multidimensional}, effectively eliminating the necessity to specify $ m_x $ and $ m_u $:
\begin{subequations}
\label{equ:rmpc_opt_problem}
\begin{alignat}{2}
& \mathcal{S}_N(x_k,y^d_k) & & = \!\min \ J_N(x_k,y^{d}_k;v(\cdot|k),y^{\zeta},x^{\zeta},v^{\zeta}) \nonumber \\
& \text{subject to} & & x(0|k) = x_k, \nonumber \\
\label{equ:dyn_pred}
& & & x_{k+1|k} = f(x_{k+1|k}, v_{k+1|k}), \\
\label{equ:tightened_const}
& & & g_{j,\kappa}(x_{k+1|k},v_{k+1|k}) + c_\eta \frac{1-\rho_d^k}{1-\rho_d}\overline{w}_d \leq 0, \\
\label{equ:new_constraints}
& & & x^s = f(x^\zeta,v^\zeta), \quad y^\zeta = o_\kappa(x^\zeta,v^\zeta), \\
\label{equ:constr_kerm_set_2}\\
& & & x_{L|k} \in \mathcal{X}_{\mathrm{f}}(x^\zeta,\alpha), \\ 
& & & k = 0,...,L-1, \quad \eta = 1,...,n, \nonumber
\end{alignat}
\end{subequations}
with the objective function
\begin{equation}
\label{eq:obj_fun}
\!\begin{aligned}
    & J(x_k,y^d_k;v(\cdot|k),x^\zeta,v^\zeta) \\
    := & \sum_{k=0}^{N-1} \left(||x_{k+1|k}-x^\zeta||_\mathbb{Q}^2 + ||v(k+1|k)-v^\zeta||_\mathbb{R}^2\right),
\end{aligned}
\end{equation}
$\mathbb{Q},\mathbb{R} \succ 0$.
The terminal set is given as
\begin{equation}
\label{equ:constr_kerm_set_3}
     \mathcal{V}(x^\zeta,\alpha) := \{x \in \mathbb{R}^n | \sqrt{\mathcal{V}_\delta(x,x^{\zeta})} + \frac{1-\rho_{d}^N}{1-\rho_{d}}\overline{w}_{d} \leq \alpha \}.
\end{equation}

The optimization problem~\eqref{equ:rmpc_opt_problem} is solved at time $k$ with the initial state $x_k$. 
The optimal input sequence is denoted by $v^*(\cdot|k)$ with the control law denoted as $\boldsymbol{\mathcal{L}}(x_k,y_k^{d})=v^*(0|k)$.
The predictions over the horizon $ N $ are carried out with respect to the nominal system description in~\eqref{equ:dyn_pred}.
\begin{algorithm}[H]
    \caption{Self-adaptive Deep MPC Algorithm}
    \label{alg:selfdmpc_v3gaijin}
    \begin{algorithmic}
        \State \textbf{given} Initiate estimate $\hat{\mathcal{V}}$ of $\mathcal{V}^*$
        \State Sample an initial state $x_k$
        \State $k = 0$
        \While{$k < \mathcal{L} $} 
          \State Compute $u_{\hat{\pi}}(k)$ according to Eq.~\eqref{eq:obj_fun} 
          \State Execute $u_{\hat{\pi}}(k)$, obtain a cost $c(k)$
          \State Add the transition tuple to a buffer
          \State $k = k + 1$
        \EndWhile
        \State Using the buffer, update $\hat{\mathcal{V}}$ (Eq.~\eqref{eq:obj_fun})
    \end{algorithmic}
\end{algorithm}

\section{Experiment Results}

Through the subsequent simulation experiments, we aim to validate the theoretical insights and intuitive conjectures previously discussed~\cite{liu2024infrared,zou2022unified}. Specifically, our objectives are to demonstrate the algorithm's efficacy in managing a sparse binary reward structure~\cite{cao2017structurally}, to establish that incorporating the running cost facilitates superior convergence, and to underscore the critical significance of the Model Predictive Control (MPC) time horizon during the learning process~\cite{lyu2022study}.

We did 6 experiments based on the UR5 robot arm in the MuJoCo~\cite{todorov2012mujoco} simulation platform, following 6 different paths in different motion planning tasks. Fig.\ref{results} shows all the tracking errors for each scenario. From Fig.\ref{sub_a} to Fig.\ref{sub_f}, the results show that our method shows strong robustness and the error will converge to zero with time growing on.

\begin{figure}[!t]
\centering
\begin{center}
\includegraphics[width=0.85\columnwidth]{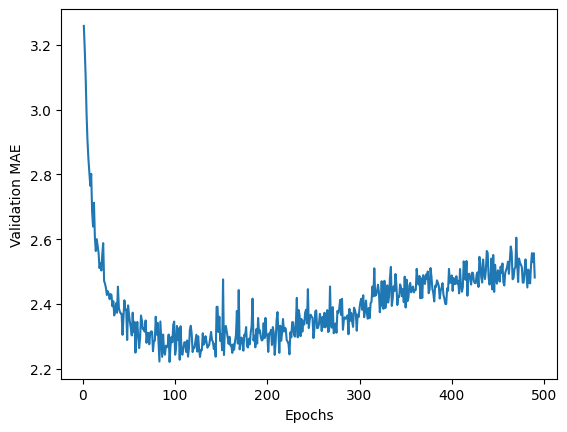}
\end{center}
 \caption{Validation mean absolute error (MAE) during the training process.}
 \label{fig3}
 \vspace{-4mm}
\end{figure}

The training process entailed utilizing images distributed uniformly across each class, with the training regimen employing batches comprising 10 samplings each. The model's performance evaluation was conducted by calculating cross-entropy loss and assessing accuracy for each batch over successive epochs, as depicted in Fig.~\ref{fig3}.

\begin{figure}[!t]
\centering
\begin{center}
\includegraphics[width=0.9\columnwidth]{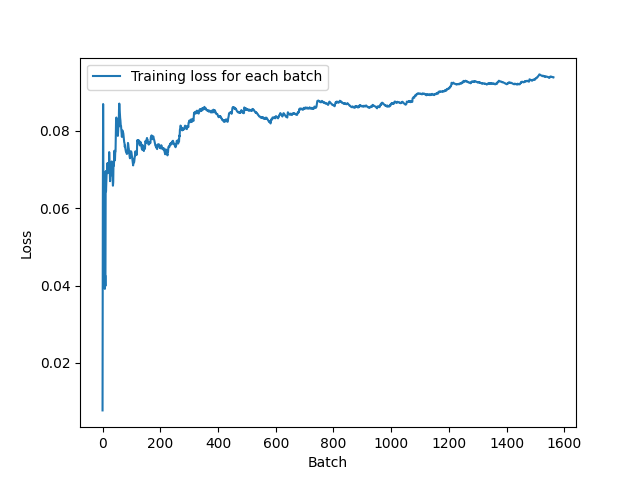}
\end{center}
 \caption{Training losses for each batch per epoch.}
 \label{fig4}
 \vspace{-4mm}
\end{figure}

Figure~\ref{fig4} illustrates the efficacy of incorporating regularization techniques during the training of our model: the validation accuracy remains constrained throughout the training process.

\section{Conclusion}
This paper introduced a self-adaptive deep MPC algorithm, which builds upon prior research integrating trajectory optimization with global value function learning. Initially, we demonstrate that leveraging a value function within the heuristic framework yields a temporally global optimal solution. Furthermore, we illustrate that employing the running cost enhances the convergence of value function estimation through an importance sampling scheme and effectively accounts for system uncertainties.

We intend to expand the value function for future research to encapsulate additional information. For instance, conditioning the value function on a local robot-centric map could enable decision-making in dynamic environments, facilitating obstacle avoidance.

\bibliographystyle{IEEEtran}
\bibliography{main}

\end{document}